\documentclass[conference]{IEEEtran}
\IEEEoverridecommandlockouts
\usepackage{cite}
\usepackage{amsmath,amssymb,amsfonts}
\usepackage{caption}
\usepackage{subcaption}
\usepackage{algorithmic}
\usepackage{graphicx}
\usepackage{textcomp}
\usepackage{makecell}
\usepackage{lipsum}
\usepackage{siunitx}
\usepackage{booktabs}
\usepackage{url}
\usepackage{hyperref}
\usepackage{xcolor}
\def\BibTeX{{\rm B\kern-.05em{\sc i\kern-.025em b}\kern-.08em
    T\kern-.1667em\lower.7ex\hbox{E}\kern-.125emX}}

\makeatletter
\newcommand{\linebreakand}{%
  \end{@IEEEauthorhalign}
  \hfill\mbox{}\par
  \mbox{}\hfill\begin{@IEEEauthorhalign}
}
\makeatother

\title{Improving Chest X-Ray Classification by RNN-based Patient Monitoring}

\author{
\IEEEauthorblockN{1\textsuperscript{st} David Biesner}
\IEEEauthorblockA{\textit{Fraunhofer IAIS and University of Bonn} \\
Sankt Augustin and Bonn, Germany \\
david.biesner@iais.fraunhofer.de}
\and
\IEEEauthorblockN{2\textsuperscript{nd} Helen Schneider}
\IEEEauthorblockA{\textit{Fraunhofer IAIS} \\
Sankt Augustin, Germany}
\and
\IEEEauthorblockN{3\textsuperscript{rd} Benjamin Wulff}
\IEEEauthorblockA{\textit{Fraunhofer IAIS} \\
Sankt Augustin, Germany}
\linebreakand
\IEEEauthorblockN{4\textsuperscript{th} Ulrike Attenberger}
\IEEEauthorblockA{\textit{University Hospital Bonn} \\
Bonn, Germany}
\and
\IEEEauthorblockN{5\textsuperscript{th} Rafet Sifa}
\IEEEauthorblockA{\textit{Fraunhofer IAIS} \\
Sankt Augustin, Germany}
}

\begin{document}
%
\maketitle
\begin{abstract}
Chest X-Ray imaging is one of the most common radiological tools for detection of various pathologies related to the chest area and lung function.
In a clinical setting, automated assessment of chest radiographs has the potential of assisting physicians in their decision making process and
optimize clinical workflows, for example by prioritizing emergency patients.

Most work analyzing the potential of machine learning models to classify chest X-ray images focuses on vision methods processing and predicting pathologies for one image at a time.
However, many patients undergo such a procedure multiple times during course of a treatment or during a single hospital stay.
The patient history, that is previous images and especially the corresponding diagnosis contain useful information that can aid a classification system in its prediction.

In this study, we analyze how information about diagnosis can improve CNN-based image classification models by constructing a novel dataset from the well studied CheXpert dataset of chest X-rays.
We show that a model trained on additional patient history information outperforms a model trained without the information by a significant margin.

We provide code to replicate the dataset creation and model training.\footnote{To be published in proceedings of IEEE International Conference on Machine Learning Applications IEEE ICMLA 2022.}
\end{abstract}

\section{Introduction}
\label{sec:intro}

Chest X-ray, the most common image examination method in the world,
is used by radiologists to aid in the diagnosis of a wide variety of life-threatening conditions like Pneumonia or Pneumothorax. 
The automatic analysis of chests radiographs has therefore the potential to optimize the clinical workflow, for example through clinical decision support systems or improved workflow prioritization \cite{irvin2019chexpert}.
Due to the increasing shortage of skilled professionals and the ever rising number of examinations conducted in hospitals,
improvements in clinical workflows are of major importance to ensure a consistent quality of patient care.

State-of-the-art methods for the detection of thoracic diseases in chest X-rays focus on the evaluation of a single, usually frontal, chest X-ray.  
However, due to the time-, cost- and radiation-efficiency of X-ray, multiple image data are often taken of a patient \cite{KERMANY20181122}.
Once a new image is taken, the previous images still contain valuable information that a clinical physician can use to monitor the development of certain pathologies and better classify the new information.

In addition to the sequential image data, a report text is available for each retrospective image, which contains the classification information of the image. 
While these unstructured reports themselves are not easily integratable in an image classification pipeline,
very capable natural language processing (NLP) algorithms have be developed to automatically classify report texts and thus image data \cite{irvin2019chexpert, smit2020chexbert}.
It can therefore be assumed that machine-readable pathology information for older examinations are available. 
Within this work, we denote this combination of previous sequential image data with the corresponding findings as the patient history. 

\begin{figure}
    \centering
    \includegraphics[width=\columnwidth]{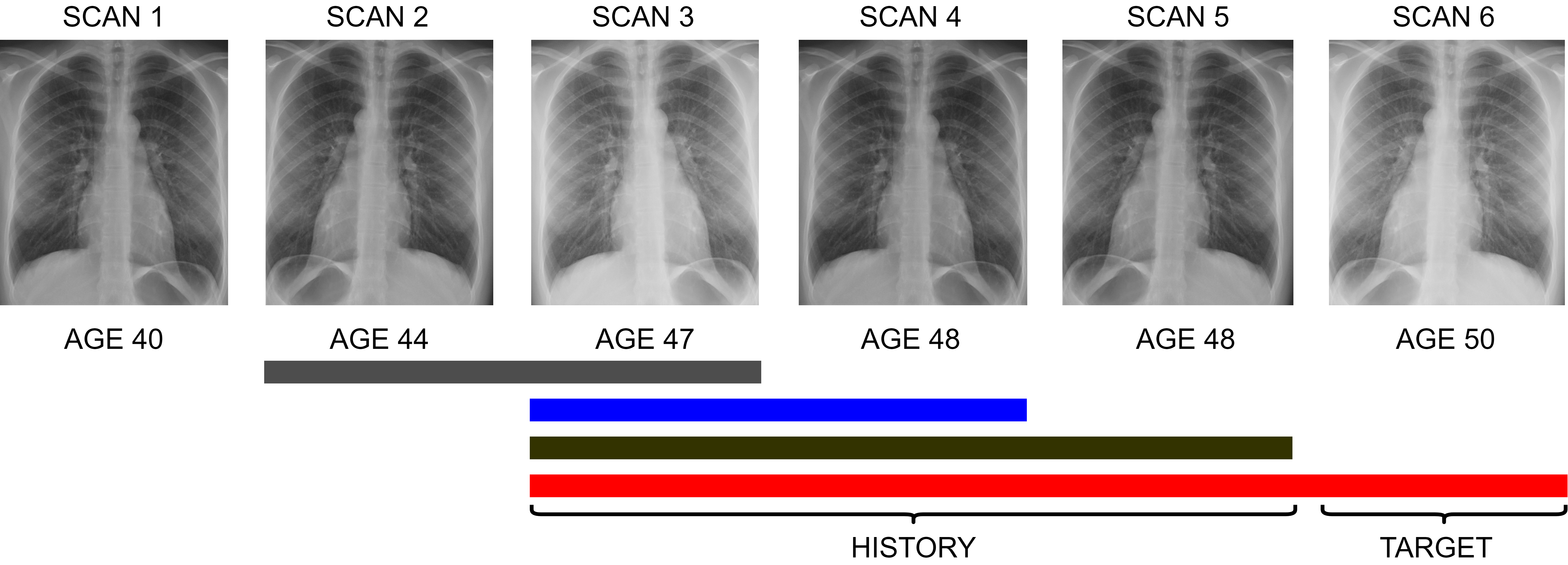}
    \caption{
    Diagram of our dataset creation process for $n_{\text{max\_age\_diff}} = 3$ and $n_{\text{min\_images}}$ = 2. 
    Here, 6 individual scans were taken of a single patient at varying ages.
    The colored bars under the images mark individual datapoints in our dataset.
    The bottom datapoint (red) contains 4 images. Scan 6 at age 50 is the target scan, scans 3 to 5 at ages 47 to 48 are the history.
    Note that scans can be part of the dataset both as target and as history.
    Additionally note that some scans from the full dataset are not present in the new patient history dataset.
    Scan 2 is not a target image in the dataset since it does not have any history images with age difference less than 3.
    Scan 1 is not part of the new dataset at all, since it does not have any history images and is not part of any history.
    Image from \cite{xray_image}.
    }
    \label{fig:images}
\end{figure}

Previous approaches do not sufficiently exploit the potential of the patient history information 
by processing one image at the time. 
Physicians also take a patient's history i.e. both older image recordings and reports texts, into account when making a diagnosis.
This leads to the motivation that the automatic analysis of chest X-rays can also be improved by the patient history. The sequential processing of the patient's image data exploits the data potential of the clinical practice during modeling.

The aim of this work is therefore to combine elements of time series analysis with image processing methods and thus to enable the processing of the entire available information.
This includes
\begin{itemize}
    \item the describtion of the curation process for generating a novel dataset of patient histories from the CheXpert dataset of chest X-ray images,
    \item an overview of model architecture and argue for the inclusion of previous expert labels in the classification process,
    \item an evaluation of the effectiveness of the proposed method against baseline methods in which we show that, when available, patient history provides valuable information to the system and
    \item an outlook into open questions for further research.
\end{itemize}

Code to replicate all experiments in this study is available at \url{https://github.com/fraunhofer-iais/rnn_patient_monitoring}.

\section{Related work}
\label{sec:related_work}
The evaluation of chest X-rays with regard to the automatic detection of diseases is an ongoing field of study. The Standford Machine Learning Group developed a comprehensive public image dataset, CheXpert, for this task using automatically generated labels \cite{irvin2019chexpert}. 
The dataset has enabled the development of various machine learning methods for the classification of chest diseases, e.g. implementation of deep neural networks, transfer learning or label smoothing methods \cite{ESANN,chexpert_labeluncertainty,chexpert_aucloss,mcdermott2020chexpert}. 

However, most of the models mentioned do not consider all the image information available for a patient. Multi-view classification models can be developed for example by taking into account the lateral scan in addition to the frontal view  \cite{multi_lateral, multi_lateral2}, improving the AUROC score of the baseline one-view model by up to 2\%.
In \cite{Sequence_img}, sequential image data analysis techniques are implemented to improve lung disease classification. The sequential aspect of the input is artificially generated by processing the same image with different networks, and does not arise from the consideration of temporally different acquisitions.

As described in the introduction, image data of the same modality with associated report texts are often available for a patient in everyday clinical practice. To the best of our knowledge this sequential image data analysis of the patient history was not considered in the evaluation of chest X-rays yet. 
We would like to fill this gap through our research and investigate whether higher performance can be achieved if the image data potential of a patient is better exploited.

Our models are based on the DenseNet \cite{DenseNet} architecture, which is a type of convolutional neural network \cite{CNN} already applied for the analysis of chest x-rays in previous studies \cite{chexpert_labeluncertainty, AnaXNet, SSCI, Covid_DenseNet}.
We model the patient history using a recurrent neural network called GRU (gated recurrent unit) \cite{GRU}, a architecture that has already been applied to a various of time series applications \cite{GRU_water, GRU_cyber, GRU_unsupervised}.

\section{Data}
\label{sec:data}

The CheXpert dataset contains a total of $\num{224316}$ chest X-ray scans from $\num{65240}$ individual patients.
There are 14 observation available relating to a variety of thoracic diseases,
parsed from the corresponding reports written by clinical radiologists.
For details on the data acquisition process see \cite{irvin2019chexpert}.

From the 14 observations, the CheXpert challenge proposes a subset of 5 pathologies as classification target due to their prevalence and clinical importance:
\begin{enumerate}
    \item Cardiomegaly, 
    \item Edema, 
    \item Consolidation, 
    \item Atelectasis and 
    \item Pleural Effusion.
\end{enumerate}
In order to keep our methods comparable to literature we restrict our model training and evaluation on the 5 pathologies as well. 
We further restrict the dataset to only frontal (as opposed to lateral) scans, which reduces the size of the dataset to \num{191027}.

Each observation in the dataset is labeled as either positive (pathology detected in the scan), negative (pathology not detected in the scan) or uncertain. The term `uncertain'
can refer both to the uncertainty of the radiologist in diagnosing and also the ambiguity of the report \cite{irvin2019chexpert}.
To deal with uncertain labels and convert the prediction problem into a binary classification 
problem, we adopt the method of \cite{chexpert_labeluncertainty}.
The uncertain labels of the pathologies `Edema', `Atelectasis' are mapped to a positive, and all uncertain labels of `Cardiomegaly', `Consolidation' and `Pleural Effusion' to a negative label. The mapping was selected to optimize performance on the hold-out hand-labeled  test set provided by CheXpert, which does not contain uncertain labels \cite{irvin2019chexpert}.

\begin{table}[]
\begin{tabular}{@{}lllllllll@{}}
\toprule
Dataset  & Images       & Datapoints       & $R_0$ & $R_1$ & $R_2$ & $R_3$ & $R_4$ \\ 
\midrule
train    & \num{105179} & \num{73340}      & 13.9  & 37.5  & 8.2   & 31.1  & 52.0  \\
valid    & \num{13270}  & \num{9299}       & 13.3  & 38.4  & 7.9   & 32.5  & 51.9  \\
test     & \num{12715}  & \num{8717}       & 12.6  & 37.6  & 8.2   & 31.2  & 52.9 \\
full    & \num{191027} & \num{191027}     & 12.3  & 33.2  & 6.8   & 31.2  & 40.4  \\
\bottomrule \\
\end{tabular}
\caption{Full statistics on our proposed datasets. Sets train, valid and test refer to our training, validation and test split for images with available patient history. These datasets contain more total images than target images (i.e. datapoints), since some images are only part of a patients history, see also Figure \ref{fig:images}.
Statistics $R_0$, \dots, $R_4$ depict the ratios of positive labels to negative labels  in the split for the labels Cardiomegaly,  Edema, Consolidation, Atelectasis and Pleural Effusion respectively. For example, of the \num{82220} target images in the training split, \num{11252} images (\num{13,7}\%) were positive for Cardiomegaly.
We additionally provide statistics for the full dataset without any restrictions on patient history (full).
}
\label{tab:dataset}
\end{table}

For each scan the dataset provides additional metadata,
like the patients sex and age.
We utilize the age information present for each scan to create a novel dataset for patient monitoring.

We sort all scans by patient ID.
For each scan, we filter the patient scans for scans with age at least $n_{\textbf{max\_age\_diff}}$ 
years less than the target scan age. We remove all scans with higher age.
We filter all scans from the dataset for which we find less than $n_{\text{min\_images}}$ previous images.

For $n_{\text{max\_age\_diff}} = 3$ and $n_{\text{min\_images}} = 1$ (i.e. scans with at least one additional image that is not more than 3 years old) we end up with a dataset of \num{82220} scans with corresponding patient history.

We split the patient monitoring dataset into a training, a validation and a test split.
We split the dataset into training ($80\%$), validation ($10\%$) and test ($10\%$) by patients, meaning scans from a single patient are only present in one of the splits.
We receive a dataset of \num{73340} target scans in the training split, \num{9299} target scans in the validation split and \num{8717} target scans in the test split.
See Table \ref{tab:dataset} for a full overview of the dataset statistics,
with additional information on the ratio of positive examples for each class.

Note that a single scan can be present multiple times in the dataset: as a target image with corresponding patient history or as part of a patient history for a later target image.
But an image can not be part of two dataset splits, since we split the dataset by patients.
See Figure \ref{fig:images} for a diagram of the dataset creation.


We preprocess each image as follows.
We first resize the image to 224x224 pixels,
the size of the ImageNet dataset \cite{imagenet}.
This is done in order to maximize the effect of using image classification models pretrained on the ImageNet dataset.
Pretraining on ImageNet has been shown to improve classification performance in downstream tasks in various domains, including medical images \cite{chexpert_labeluncertainty},
and that this size is sufficient for classification of the various pathologies present in the dataset \cite{chexpert_aucloss,chexpert_labeluncertainty}. 
Further we apply a normalization scheme as in \cite{chexpert_labeluncertainty},
which normalizes each image in the dataset to fit the mean and variance of color present in ImageNet.
Before each training step we apply a random rotation, shift and zoom to each image.
We do not apply random transformations to the images in the validation and test split.

\begin{figure}
    \centering
    \begin{subfigure}[b]{\columnwidth}
    \includegraphics[width=\columnwidth]{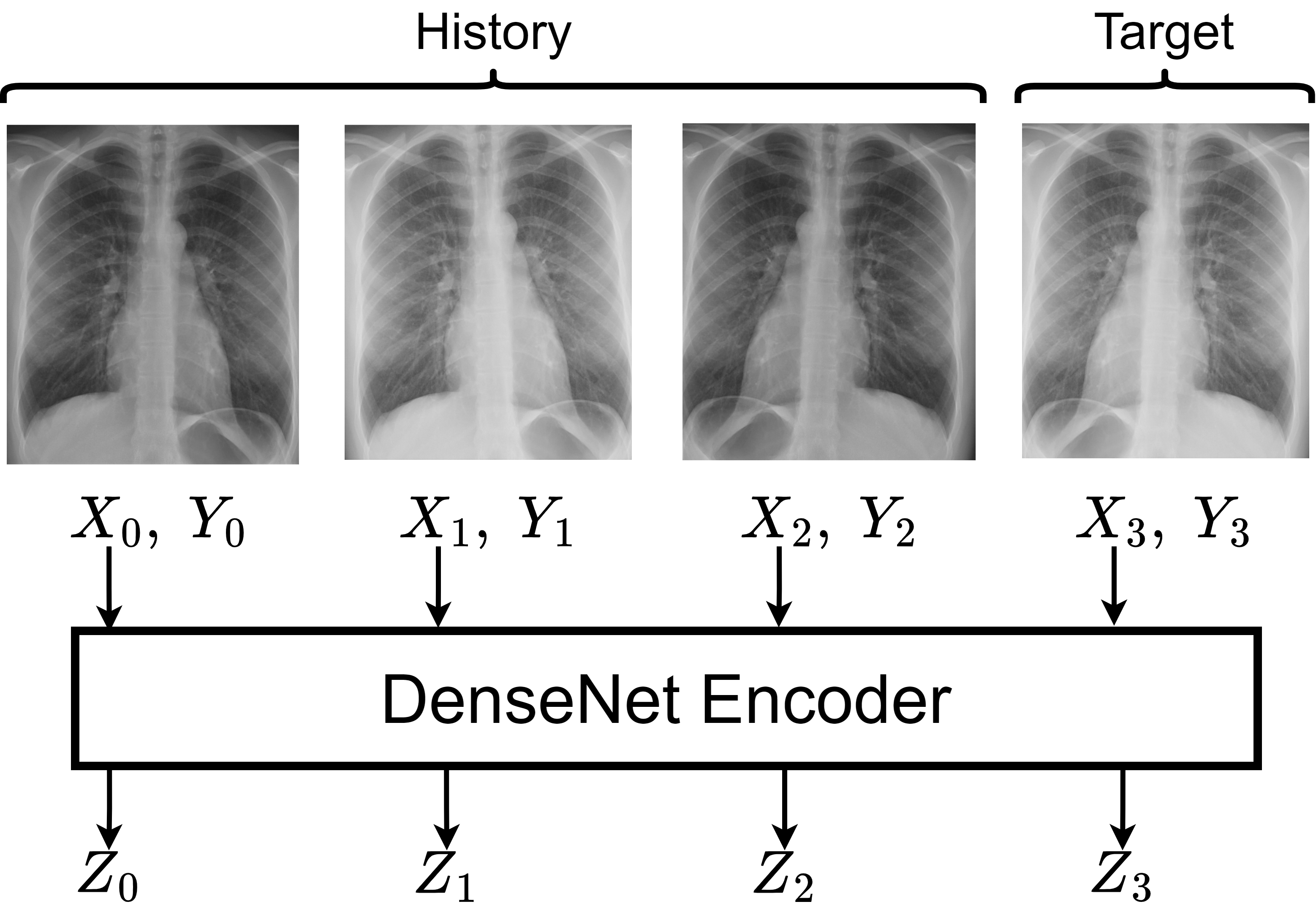}
    \caption{
    Diagram of the encoding process for one datapoint with one target image $X_3$ and three history images $X_0, X_1, X_2$
    (with corresponding labels $Y_0, \dots, Y_3$).
    The DenseNet encoder processes each $X_i$ individually and returns a latent representation $Z_i$.
    }    
    \end{subfigure}
     \hfill
     \vspace{0.1cm}
    \begin{subfigure}[b]{\columnwidth}
    \includegraphics[width=\columnwidth]{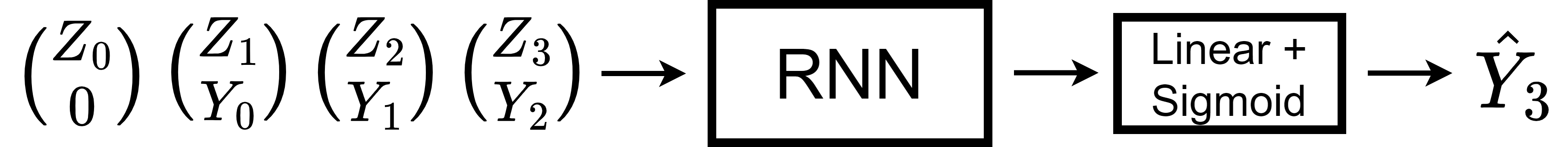}
    \caption{
    Diagram of the prediction architecture.
    We concatenate each encoded image $Z_i$ with the corresponding labels to the previous image $Y_{i-1}$.
    The first image $Z_0$ is concatenated with a vector of all zeros.
    The image-label pairs are passed through a recurrent neural net. 
    The output of the RNN is passed through a linear layer to map it onto a logit vector of size of the target label vector and a final sigmoid layer maps the logits to a probability vector $\hat{Y}_3$.
    During training, we calculate the cross binary entropy between prediction $\hat{Y}_3$ and $Y_3$.
    }    
    \end{subfigure}
    \caption{
    Schematic representation of both parts of the prediction architecture.
    }
    \label{fig:model_architecture}
\end{figure}

\begin{table*}[]
\centering
\begin{tabular}{@{}lrrrrrr@{}}
\toprule
                          & \multicolumn{3}{c}{Validation}                                                                                                    & \multicolumn{3}{c}{Test}  \\ 
                            \cmidrule(l){2-4}                                                                                                                   \cmidrule(l){5-7}
\multicolumn{1}{c}{Model} & \multicolumn{1}{c}{AUROC Macro}          & \multicolumn{1}{c}{AUROC Micro}          & \multicolumn{1}{c}{AUROC Weighted}          & \multicolumn{1}{c}{AUROC Macro}          & \multicolumn{1}{c}{AUROC Micro}          & \multicolumn{1}{c}{AUROC Weighted}    \\ 
\midrule
DenseNet Baseline         & 71.6 $\pm$ $5\mathrm{e}{-2}$             & 80.3 $\pm$ $7\mathrm{e}{-2}$             & 72.7 $\pm$ $1\mathrm{e}{-1}$                & 74.6 $\pm$ $5\mathrm{e}{-2}$             & 81.1 $\pm$ $3\mathrm{e}{-2}$             & 75.5 $\pm$ $5\mathrm{e}{-2}$          \\
RNN Image                 & 68.8 $\pm$ $1\mathrm{e}{-1}$             & 79.5 $\pm$ $7\mathrm{e}{-2}$             & 70.6 $\pm$ $7\mathrm{e}{-2}$                & 69.6 $\pm$ $7\mathrm{e}{-2}$             & 79.7 $\pm$ $1\mathrm{e}{-1}$             & 71.0 $\pm$ $1\mathrm{e}{-1}$          \\
RNN Image + Label         & \textbf{73.8 $\pm$ $8\mathrm{e}{-2}$}    & \textbf{82.1 $\pm$ $5\mathrm{e}{-2}$}    & \textbf{75.3 $\pm$ $1\mathrm{e}{-1}$}       & \textbf{74.8 $\pm$ $1\mathrm{e}{-1}$}    & \textbf{82.5 $\pm$ $6\mathrm{e}{-2}$}    & \textbf{75.7 $\pm$ $1\mathrm{e}{-1}$} \\
RNN Label                 & 72.4 $\pm$ $8\mathrm{e}{-2}$              & 81.6 $\pm$ $5\mathrm{e}{-2}$            & 73.7 $\pm$ $8\mathrm{e}{-2}$                & 73.6 $\pm$ $8\mathrm{e}{-2}$             & 82.1 $\pm$ $8\mathrm{e}{-2}$             & 74.5 $\pm$ $7\mathrm{e}{-2}$          \\ 
\bottomrule
\end{tabular}
\caption{Evaluation of the proposed RNN-based methods against the baseline DenseNet method.
We evaluate on the validation and test split of our novel patient monitoring dataset.
We compute the AUROC score and display the score over all 5 classes, as macro average, micro average or weighted average.
For each split, we apply 10x bootstrapping to receive a statistically robust performance metric. We provide the mean score over all bootstrapped subsets and the standard deviation.
}
\label{tab:eval}
\end{table*}

\section{Models}
\label{sec:models}

The base architecture for all experiments in this study is DenseNet \cite{DenseNet}.
Based on previous work \cite{chexpert_aucloss,chexpert_labeluncertainty,mcdermott2020chexpert,ESANN} we use a single DenseNet model,
pretrained on ImageNet, with a linear classifier for all 5 labels, as a baseline.

The patient monitoring model we propose in this study employs a DenseNet architecture as well.
For each image with associated patient history images we encode each image into a latent vector
using the same pretrained DenseNet as the baseline model.
The latent vectors are then passed through a bidirectional recurrent neural net (RNN).
The output of the last layer of the RNN is then passed to a linear classification layer.
Finally the logit outputs of the classification layer are passed through a sigmoid layer,
outputting probabilities for each label.
As RNN architecture we use a GRU (gated recurrent unit \cite{GRU}),
which is similar to an LSTM (long short term memory)
but contains less parameters.
Our experiments have shown that the lower parameter count makes the model less prone to overfitting and the rather short sequences of images do not necessitate the specific long term memory of LSTMs.

In order to make use of information given by radiologists when examining the previous images,
we enrich the RNN with the label information.
For this, we append a binary vector over all labels of the corresponding image to each encoded image passed to the RNN.
See Figure \ref{fig:model_architecture} for a more detailed description of the proposed model.

We argue that this method is valid when evaluation on a validation set and when predicting labels in a clinical setting.
Each X-ray scan conducted in a hospital will eventually be examined by an expert radiologists.
Methods such as ours for automatic pathology prediction only aide the clinician in their decision making process.
Therefore all previous scans available at the hospital will have the relevant pathology information present.
We can therefore assume that our system will have access to this information during inference time.

While most examinations of chest X-rays only result in written reports by radiologists, not structured label information,
there has been much work conducted on the automatic extraction of labels from free text reports.
The dataset considered in this study, CheXpert \cite{irvin2019chexpert}, itself provides labels extracted by rule-based methods from free text reports.
Further studies \cite{smit2020chexbert} have examined the use of sophisticated transformer-based language models \cite{bert} for label extraction and further improved the accuracy.
Application of such algorithms to automatically analyze written reports and use extracted information to enhance the image classification system is feasible in a clinical setting.

In order to examine whether the decision making process of the model is only dependent on the previous labels, we train an additional RNN-based model only on the label information, we therefore ignore the actual images and pass the binary label vector directly to the RNN.

\section{Training and Evaluation}
\label{sec:results}

\subsection{Training Details}

For this evaluation we train 4 distinct models:
\begin{itemize}
    \item \emph{DenseNet Baseline}: The baseline DenseNet model used in various previous studies \cite{chexpert_labeluncertainty,irvin2019chexpert},
    \item \emph{RNN Image}: The proposed RNN-based model with only the encoded images passed through the recurrent neural net,
    \item \emph{RNN Image + Label}: The proposed RNN-based model with both the encoded images and corresponding labels passed through the recurrent neural net, and
    \item \emph{RNN Label}: The proposed RNN-based model with only the labels passed through the recurrent neural net.
\end{itemize}

We train each model on the training split of our patient monitoring dataset until convergence of the 
binary-cross-entropy loss and choose the model with the lowest validation loss for evaluation on the test set.
We train the models for a maximum of 30 epochs with a learning rate of $1\mathrm{e}{-3}$ and the  reduce-on-plateau learning rate scheduler using the Adam \cite{adam} optimizer.

\subsection{Evaluation}

We evaluate all trained models against the validation and test split of our patient monitoring dataset.
To receive a statistically robust estimate the model performance, we apply Poisson bootstrapping \cite{bootstrapping}
to resample 10 validation and test splits.
We evaluate the metrics over all bootstrapped datasets and compare mean and standard deviation.

Our models output a probability vector over all labels for each image.
The probability denotes the confidence of the model that a certain pathology is visible in the image.
In practice, one would apply a threshold to classify pathologies as present or not present, depending on the probability output of the model.
However, choosing a threshold introduces bias towards one type of classification error,
and therefore thresholds must be determined with a certain use case in mind.
Metrics such as Precision, Recall, Accuracy and F1-Score require a certain classification threshold to be set.
We therefore consider a threshold independent metric in this study, the area under the receiver operating characteristic curve (AUROC \cite{AUROC}).
A perfect AUROC score is achieved if every positive sample is assigned a higher probability than every negative sample,
i.e. if a threshold with 100\% accuracy is possible.
The AUROC score then denotes the ratio of correctly sorted examples in the dataset.

We average the AUROC score over the 5 classes using three methods:
\begin{itemize}
    \item macro averaging: unbiased average of all five scores,
    \item weighted averaging: average of all five scores, weighted by class support, i.e. number of positive examples in each class, and
    \item micro averaging: global score over all samples and classes.
\end{itemize}

We evaluate our proposed methods against the DenseNet baseline in Table \ref{tab:eval}.
Considering the AUROC metrics on the test set,
we see that the RNN-model with both image and label information (\emph{RNN Image + Label}) scores higher than
the baseline for all averaging methods.
This holds as statistically significant considering the 10-factor bootstrapping and the displayed standard deviation.
The most significant difference shows in the micro-averaged AUROC score on the test set,
in which the proposed model scores $1.4$ percentage points higher than the baseline.

We do not see this full effect when we restrict the information the RNN model receives to only the encoded images or even the label vectors.
The \emph{RNN Image} model scores lower than both the baseline and the \emph{RNN Image + Label} model in all evaluations.
The \emph{RNN Label} model scores lower than the \emph{RNN Image + Label} model in all evaluations, 
and only slightly higher than the baseline in some.
Both RNN-based models with only partial patient history information are not able fully utilize the additional data and reliably improve on the baseline.
In order to take full advantage of the patient history information available in the novel dataset,
we must present both the encoded images and the previous label vectors to the model.


\section{Conclusion and Outlook}

In this study we proposed a novel method for classification of chest X-ray images
and argued how taking into account information from the patient history can aide image classification systems
in the prediction of clinical pathologies found in chest radiographs.

The proposed model consists a recurrent neural net, which processes the sequence of encoded patient history images with corresponding pathology labels.
We determine that the inclusion of real labels from the dataset is justifiable in this setup,
since in a clinical setting previous X-ray images will always have been examined by an expert radiologist.

To test the method we implemented a novel dataset based on the well-known CheXpert dataset of chest X-rays.
The method consistently improves the classification performance if such patient history is available for a given image.
We release the code for training and dataset reproduction,
such that more future research can investigate the effect of inclusion of patient history in the image classification pipeline.

Open research questions include:
\begin{itemize}
    \item Does the inclusion of patient history benefit image models trained on other chest X-ray datasets or other medical imaging datasets?
    \item We see an improvement of the baseline method when increasing the number of training images to the full dataset.
    How does the performance of the proposed model improve when trained on more data?
    \item Are more sophisticated sequence models (like transformer architectures) better at parsing information from the encoded sequences or does the increase in parameter count only work towards overfitting the model?
    \item Can the direct comparison of patient history image data further increase the performance of the model?
\end{itemize}

We plan to address these questions in future work and hope to implement our methods in practice applications soon.

\bibliographystyle{IEEEbib}
\bibliography{refs}

\end{document}